\documentclass[11pt,a4paper]{article}
\usepackage{authblk}
\usepackage[hyperref]{emnlp2020}
\usepackage{times}
\usepackage{float}
\usepackage{latexsym}
\usepackage{graphicx}

\usepackage[utf8]{inputenc}	

\usepackage{pbox}
\usepackage{multirow}
\usepackage{cleveref}

\usepackage{microtype}
\usepackage{csquotes}
\aclfinalcopy 
\def\aclpaperid{2391} 

\title{RussianSuperGLUE: A Russian Language Understanding Evaluation Benchmark}

\author[1,2]{Tatiana Shavrina}
\author[1]{Alena Fenogenova}
\author[1,3]{Anton Emelyanov}
\author[1]{Denis Shevelev}
\author[2]{Ekaterina Artemova}
\author[4]{\\Valentin Malykh}
\author[1,2]{Vladislav Mikhailov}
\author[1,2]{Maria Tikhonova}
\author[1]{Andrey Chertok}
\author[1]{Andrey Evlampiev}

\affil[1]{Sberbank / Moscow, Russia}
\affil[2]{National Research University  Higher School of Economics / Moscow, Russia}
\affil[3]{Moscow Institute of Physics and Technology / Moscow, Russia}
\affil[4]{Huawei  Noah’s Ark lab/ Moscow, Russia}

\date{}

\begin{document}
\maketitle
\begin{abstract}
In this paper, we introduce an advanced Russian general language understanding evaluation benchmark -- RussianGLUE.  \\
Recent advances in the field of universal language models and transformers require the development of a methodology for their broad diagnostics and testing for general intellectual skills - detection of natural language inference, commonsense reasoning, ability to perform simple logical operations regardless of text subject or lexicon. For the first time, a benchmark of nine tasks, collected and organized analogically to the SuperGLUE methodology~\cite{wang2019superglue}, was developed from scratch for the Russian language. We provide baselines, human level evaluation,  an open-source framework for evaluating models 
 and an overall leaderboard of transformer models for the Russian language.\\
Besides, we present the first results of comparing multilingual models in the adapted diagnostic test set and offer the first steps to further expanding or assessing state-of-the-art models independently of language.

\end{abstract}

\section{Introduction} \label{sec:intro}

With the development of technologies for text processing and then deep learning methods for obtaining better text representation, language models went through the increasingly advanced stages of natural language modelling. 

Modern scientific methodology is beginning to gradually explore universal transformers as an independent object of study - furthermore, such models show the ability to extract causal relationships in texts (natural language inference), common sense and world knowledge and logic (textual entailment), to generate coherent and correct texts. An actively developing field of model interpretation develops testing procedures comparing their performance to a human level and even the ability to reproduce some mechanisms of human brain functions.

NLP is gradually absorbing all the new areas responsible for the mechanisms of thinking and the theory of artificial intelligence.

Benchmark approaches are being developed, testing general intellectual ``abilities'' in a text format, including complex input content, but having a simple output format. Most of these benchmarks (for more details see Section~\ref{sec:related_work}) make the development of machine intelligence anglo-centric, while other, less widespread languages, in particular Russian, have other characteristic linguistic categories to be tested.

In this paper, we expand the linguistic diversity of the testing methodology and present the first benchmark for evaluating universal language models and transformers for the Russian language, together with a portable methodology for collecting and filtering the data for other languages.

The contribution of RussianGLUE is two-fold. First, it provides nine novel datasets for the Russian language covering a wide scope of NLU tasks. The choice of the tasks are justified by the design of prior NLU benchmarks \cite{wang2018glue,wang2019superglue}. Second, we evaluate two widely used deep models to establish baselines. 

The remainder is structured as follows. We overview multiple prior works on developing NLU benchmarks, including those designed for languages other than English, in Section~\ref{sec:related_work}. Section~\ref{sec:tasks} lists the tasks and novel datasets, proposed for the Russian NLU. Section~\ref{sec:experiments} presents with the baselines, established for the tasks, including a human level baseline. We overview compare achieved results in Section~\ref{table:results} to the current state of English NLU. We discuss future work directions and emphasize the importance of NLU benchmarks for languages other than English in Section~\ref{sec:discussion}. Section~\ref{sec:conclusion} concludes.  

\section{Related Work} \label{sec:related_work}

Several benchmarks have been developed to evaluate and analyze word and sentence embeddings over the past few years. 


SentEval \cite{conneau2018senteval} is one of the first frameworks intended to evaluate the quality of sentence embeddings. A twofold set of transfer tasks is used to assess the generalization power of sentence embedding models. The transfer tasks comprise downstream tasks, in which the sentence embedding is used as a feature vector, and probing tasks, which are aimed to evaluate the capability of sentence embeddings to encode linguistic properties. The choice of the downstream tasks is limited to sentiment classification, natural language inference, paraphrase detection and image captioning tasks. The probing tasks are meant to analyse morphological, syntactical and semantical information encoded in sentence embeddings.  

The General Language Understanding Evaluation (GLUE) \cite{wang2018glue} benchmark is a collection of tools for evaluating the performance of language models across a diverse set of existing natural language understanding (NLU) tasks, adopted from different sources. These tasks are divided into two parts: single sentence classification tasks and sentence pair classifications tasks subdivided further into similarity and inference tasks. GLUE also includes a hand-crafted diagnostic test, which probes for complex linguistic phenomena, such as the ability of the model to express lexical semantics and predicate-argument structure, to pose logical apparatus and knowledge representation. GLUE  is recognized as a de-facto standard benchmark to evaluate transformer-derived language models. Last but not least GLUE informs on human baselines for the tasks, so that not only submitted models are compared to the baseline, but also to the human performance. The SuperGLUE \cite{wang2019superglue} follows GLUE paradigm for language model evaluation based on NLU tasks, providing with more complex tasks, of which some require reasoning capabilities and some are aimed at detecting ethical biases. A few recent projects reveal that GLUE tasks may be not sophisticated enough and do not require much tasks-specific linguistic knowledge \cite{kovaleva2019revealing,warstadt2019investigating}. Thus SuperGLUE benchmark, being more challenging,  becomes much more preferable for evaluation of language models. 

decaNLP \cite{McCann2018decaNLP} widens the scope for language model evaluation by introducing ten disparate natural language tasks. These tasks comprise not only text classification problems, but sequence tagging and sequence transformation problems. The latter include machine translation and text summarization, while the former include semantic parsing and semantic role labelling. Although decaNLP along with the associated research direction focuses on multi-task learning as a form of question answering, it supports zero-shot evaluation. 

To evaluate models for languages other than English, several monolingual benchmarks were developed, such as FLUE \cite{le2019flaubert} and CLUE \cite{CLUE}, being French and Chinese versions of GLUE. These benchmarks include a variety of tasks, ranging from part-of-speech tagging and syntax parsing to machine reading comprehension and natural language inference. 

To the best of our knowledge, LINSPECTOR \cite{eichler-etal-2019-linspector} is a first multi-lingual benchmark for evaluating the performance of language models. LINSPECTOR offers 22 probing tasks to analyse for a single linguistic feature such as case marking, gender, person, or tense for 52 languages. A part of these 22 probing tasks are static, i.e. are aimed at evaluation of word embeddings, and the rest are contextual and should be used to evaluate language models. Released in early 2020 two multilingual benchmarks, \cite{liang2020xglue} and  XTREME \cite{hu2020xtreme}, aim at evaluation of cross-lingual models.  
XGLUE includes 11 tasks, which cover both language understanding and language generation problems, for 19 languages. XGLUE provides with several multilingual and bilingual corpora that allow of cross-lingual model training. As for the Russian language, XGLUE provides with four datasets for POS tagging, a part of XNLI \cite{conneau2018xnli} and two datasets, crawled from commercial news website, used for news classification and news headline generation.  XTREME consists of nine tasks which cover classification, sequence labelling, question answering and retrieval problems for 40 languages. Almost a half of the datasets were translated from English to the target languages with the help of professional translators. XTREME offers for the Russian language five datasets, including NER and two question-answering datasets. Both XGLUE and XTREME offer tasks that are much simpler than SuperGLUE and are aimed at evaluation of cross-lingual models rather than at comparison of mono-lingual models in similar setups. Thus the need  for novel datasets targeted at mono-lingual model evaluation for languages other than English is still not eliminated. 

\section{RussianGLUE Overview} \label{sec:overview}
We have intenooed to have the same task set in the framework as one in the SuperGLUE. There is no one-to-one mapping, but the corpora we use could be considered close to the specified tasks in the SuperGLUE framework.

We divided the tasks into six groups, covering the general diagnostics of language models and different core tasks: common sense understanding, natural language inference, reasoning, machine reading and world knowledge.     

\subsection{Tasks} 
\label{sec:tasks}

\begin{figure}[!h]
    \centering
\includegraphics[width=\linewidth]{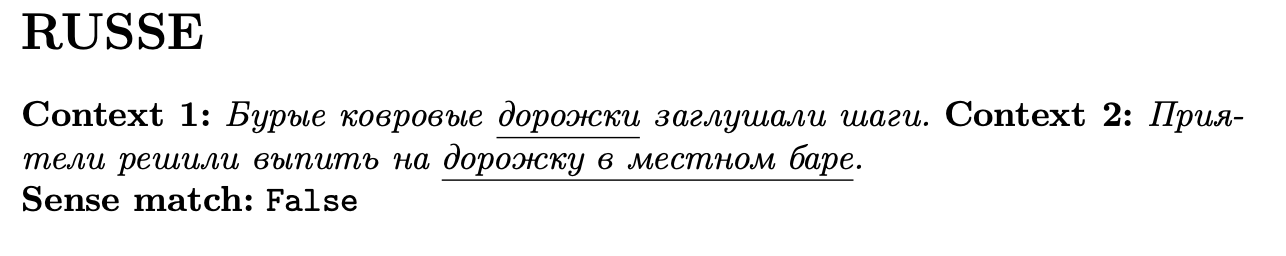}
\caption{A sample from RUSSE dataset.}
\label{fig:sample-russe}
\end{figure}

\begin{figure}[ht!]
    \centering
   \includegraphics[width=\linewidth]{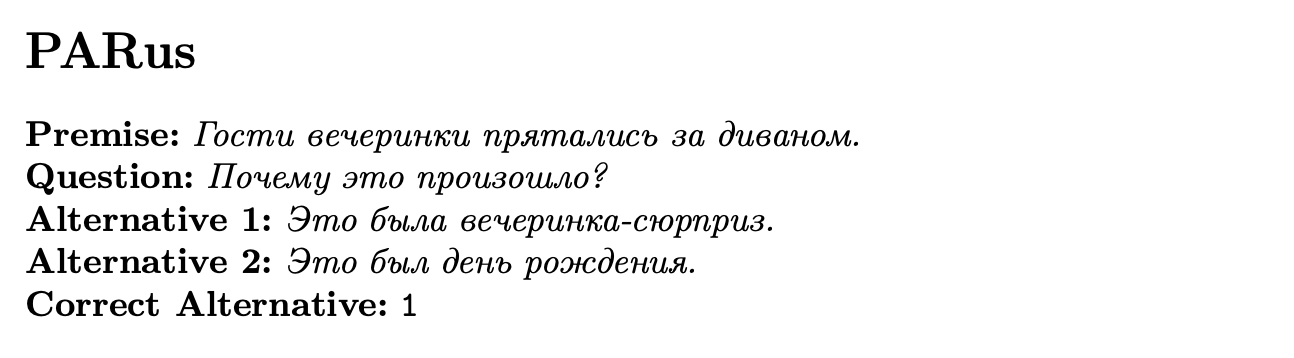}
   \caption{A sample from PARus dataset.}
   \label{fig:sample-parus}
\end{figure}

\begin{figure}[h!]
    \centering
\includegraphics[width=\linewidth]{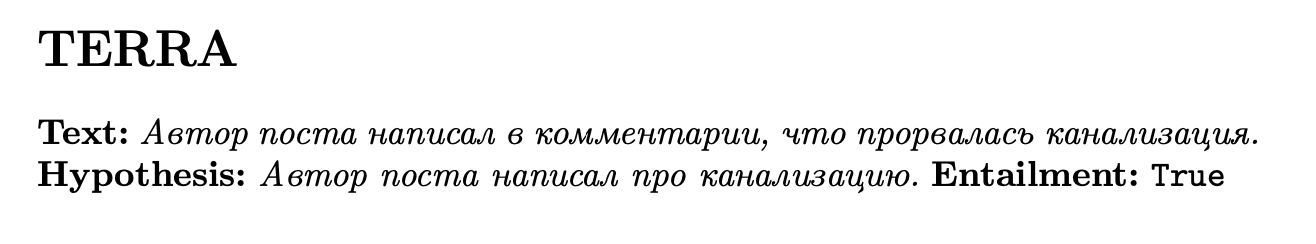}
\caption{A sample from TERRa dataset.}
\label{fig:sample-terra}
\end{figure}

\begin{figure}[h!]
    \centering
\includegraphics[width=\linewidth]{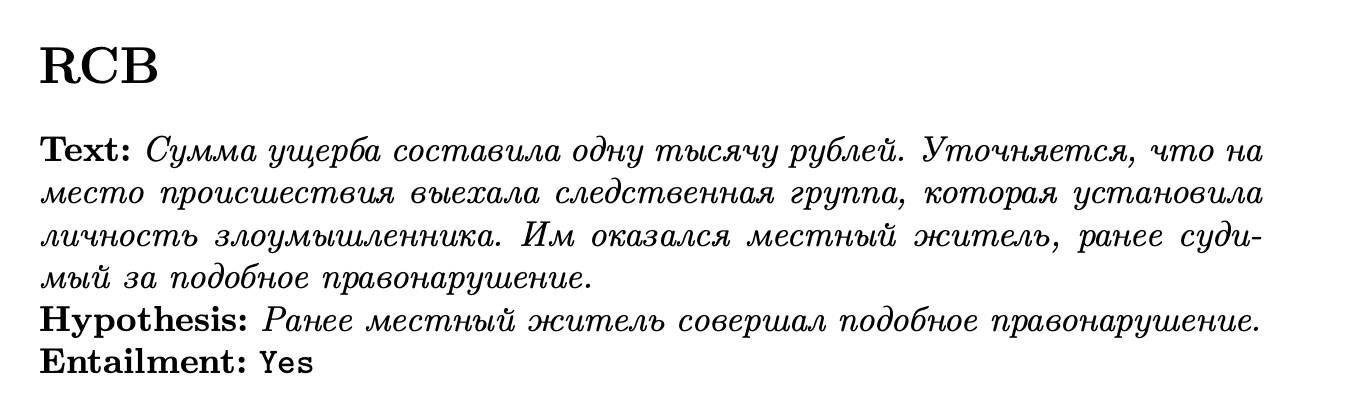}
\caption{A sample from RCB dataset.}
\label{fig:sample-rcb}
\end{figure}

\begin{figure}[h!]
    \centering
    \includegraphics[width=\linewidth]{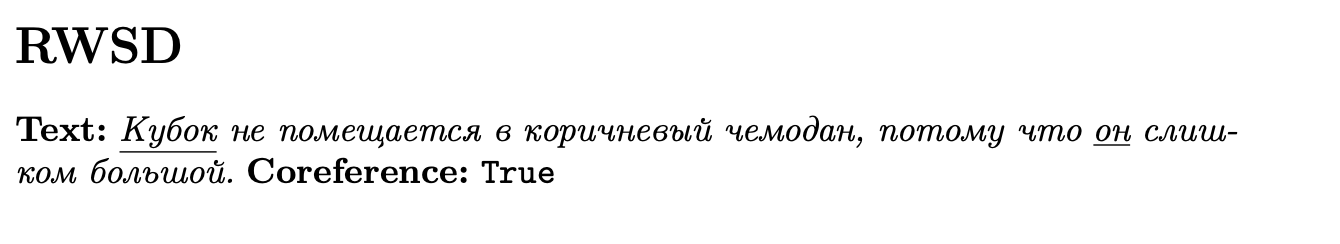}
    \caption{A sample from RWSD dataset.}
    \label{fig:sample-rwsd}
\end{figure}

\begin{figure}[h!]
    \centering
\includegraphics[width=\linewidth]{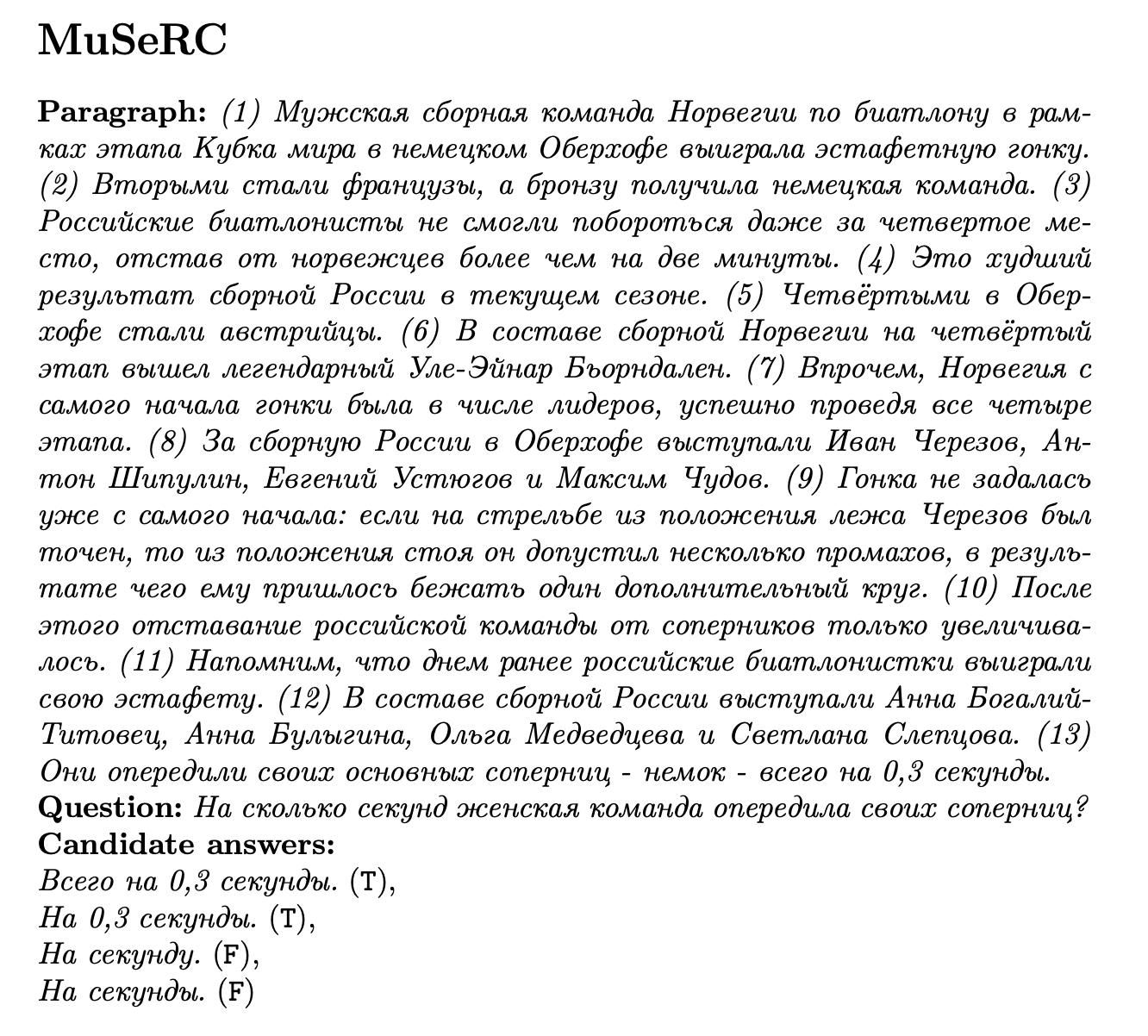}
\caption{A sample from MuSeRC dataset.}
\label{fig:sample-muserc}
\end{figure}

\begin{figure}[h!]
    \centering
   \includegraphics[width=\linewidth]{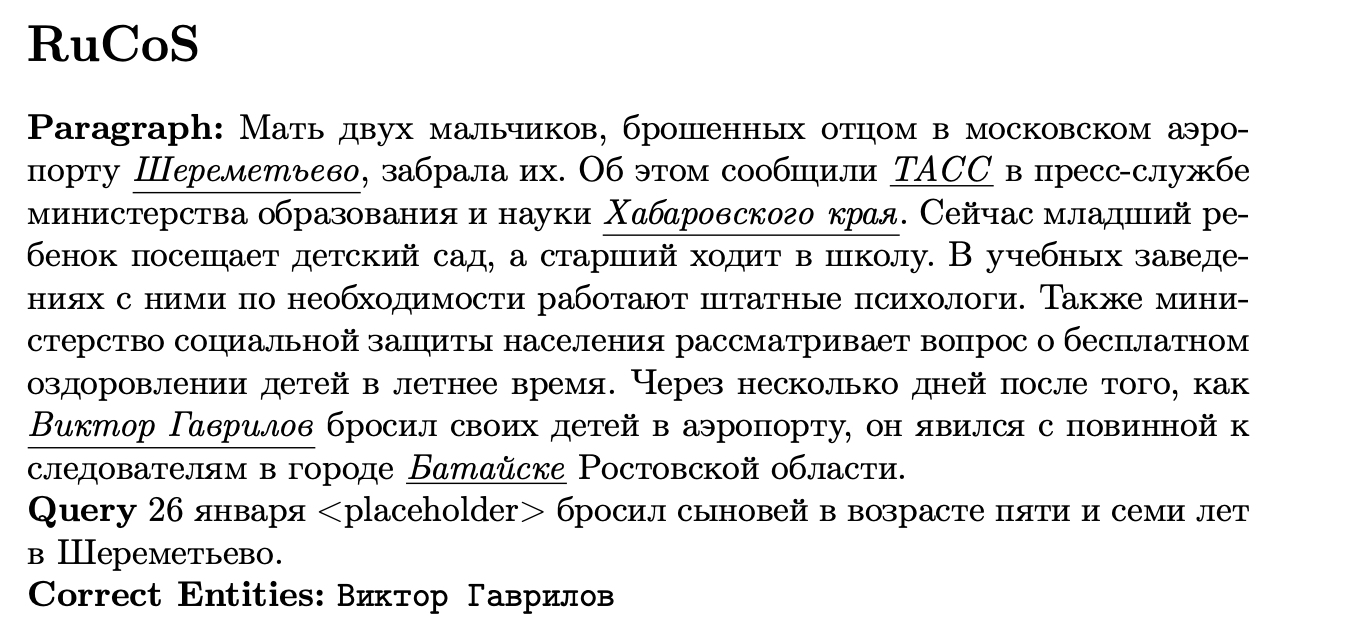}
   \caption{A sample from RuCoS dataset.}
   \label{fig:sample-rucos}
\end{figure}

The tasks description is provided below. The samples from the tasks are presented at~\cref{fig:sample-russe,fig:sample-rucos,fig:sample-rwsd,fig:sample-terra,fig:sample-muserc,fig:sample-parus,fig:sample-rcb}.

\subsubsection{Diagnostics}  \label{sec:diagnostics}

\textbf{LiDiRus}: Linguistic Diagnostic for Russian is a diagnostic dataset that covers a large volume of linguistic phenomena, while allowing you to evaluate information systems on a simple test of textual entailment recognition. This dataset was translated from English to Russian with the help of professional translators and linguists to ensure that the desired linguistic phenomena remain. This dataset corresponds to AX-b dataset in SuperGLUE benchmark. 

\subsubsection{Common Sense}

\textbf{RUSSE}: Word in context is a binary classification task, based on word sense disambiguation problem. Given two sentences and a polysemous word, which occurs in both sentences, the task is to determine, whether the word is used in the same sense in both sentences, or not. For this task we used the Russian word sense disambiguation dataset \textbf{RUSSE}  \cite{panchenko2015russe} and converted it into WiC dataset format from SuperGLUE.  

\textbf{PARus}: The choice of Plausible Alternatives for Russian language evaluation provides researchers with a tool for assessing progress in open-domain commonsense causal reasoning. Each question in PARus is composed of a premise and two alternatives, where the task is to select the alternative that more plausibly has a causal relation with the premise. The correct alternative is randomized so that the expected performance of randomly guessing is 50\%.
PARus is constructed as a translation of COPA dataset from SuperGLUE and edited by professional editors. The data split from COPA is retained.

\subsubsection{Natural Language Inference}

\textbf{TERRa}: Textual Entailment Recognition for Russian is a dataset which is devoted to capture textual entailment. The task of textual entailment has been proposed recently as a generic task that captures major semantic inference needs across many NLP applications, such as Question Answering, Information Retrieval, Information Extraction, and Text Summarization. This task requires to recognize, given two text fragments, whether the meaning of one text is entailed (can be inferred) from the other text.  The corresponding dataset in SuperGLUE is RTE, which in its place is constructed from NIST RTE challenge series corpora. To collect TERRa we filtered out the large scale Russian web-corpus, Taiga \cite{shavrina2017methodology} with a number of rules to extract suitable sentence pairs and manually corrected them. The rules had the following structures: there should be a mental verb in the first sentence and the second sentence should be attached to the first one by a subordinate conjunction. To ensure the 
literary language of the extracted sentences, we processed only news and fiction parts of Taiga and made sure, that the sentences contain only frequently used words (i.e. number instances per million, IPM is higher than 1). The word frequencies were estimated according to Russian National Corpus\footnote{\url{http://www.ruscorpora.ru/new/en/}}. 

\textbf{RCB}: The Russian Commitment Bank is a corpus of naturally occurring discourses whose final sentence contains a clause-embedding predicate under an entailment canceling operator (question, modal, negation, antecedent of conditional). Similarly to the design of TERRa dataset, we filtered out Taiga with a number of rules and manually post processed the extracted passages.  
Final labelling was conducted by three of the authors. 
This dataset corresponds to CommonBank dataset. 

\subsubsection{Reasoning}

\textbf{RWSD}: Winograd Schema task is devoted to coreference resolution in specifically designed experiment, where reference could be resolved only using the common sense. The Russian Winograd Schema Dataset (\textbf{RWSD}) is constructed as translation of the Winograd Schema Challenge\footnote{\url{https://cs.nyu.edu/faculty/davise/papers/WinogradSchemas/WS.html}}.

\subsubsection{Machine Reading}

\textbf{MuSeRC}: Russian Multi-Sentence Reading Comprehension is a reading comprehension challenge in which questions can only be answered by taking into account information from multiple sentences. The dataset is the first to study multi-sentence inference at scale, with an open-ended set of question types that requires reasoning skills. 
The task is actually a binary classification, whether the answer to the question is correct or not. Each example consists of numerated passage, question and answers.
Our dataset contains approximately 6000 questions for more than 800 paragraphs across 5 different domains, namely: 1) elementary school texts, 2) news, 3) fiction stories, 4) fairy tales, 5) brief annotations of TV series and books.
First, we have collected open sources data from different domains and automatically preprocessed them, filtered only those paragraphs that corresponds to the following parameters: 1) paragraph length 2) number of named entities 3) number of coreference relations. Afterwords we have checked the correct splitting on sentences and numerate each of them. Next, in Toloka\footnote{\url{https://toloka.yandex.ru}} we have generated the crowd sourcing task to get the following information: 1) generate questions 2) generate answers 3) check that to solve every question a human needs more than one sentence in the text.
Collecting the dataset we adhere the principles of MultiRC \cite{MultiRC2018}: a) We exclude any question that can be answered based on a single sentence from a paragraph; b) Answers are not written in the full match form in the text; c)	Answers to the questions are independent from each other. 

\textbf{RuCoS}: Russian reading comprehension with Commonsense reasoning is a large-scale dataset for machine reading comprehension requiring commonsense reasoning. The dataset construction is based on ReCoRD methodology \cite{zhang2018record}. RuCoS consists of passages and cloze-style queries automatically generated from Russian news articles, namely Lenta\footnote{\url{https://lenta.ru/}} and Deutsche Welle\footnote{\url{https://www.dw.com/ru/}}. Each sample from the dev and test sets was validated by crowd workers. The answer to each query is a text span that corresponds to one or more referents of the answer entity in the context. The answer entity may be expressed by an abbreviation, an acronym or a set of surface forms. Hence, the task requires understanding of rich inflectional morphology and lexical variability of Russian. The goal of RuCoS is to test a machine’s ability to infer the answer based on the commonsense reasoning and knowledge.

\subsubsection{World Knowledge}

{\bf DaNetQA}: This question-answering corpus follows BoolQ \cite{clark2019boolq} design: it comprises natural yes/no questions. Each question is paired with a paragraph from Wikipedia and an answer, derived from the paragraph. The task is  to take both the question as input and a paragraph and come up with a yes/no answer, i.e. to produce a binary output. DaNetQA was collected in a few steps: 1) we used crowd workers to compose candidate yes/no questions; 2) we used Google API to retrieve relevant Wikipedia pages by treating each question as a search query; 3) we queried a pre-trained BERT-based model for SQuAD \cite{kuratov2019adaptation} to extract relevant paragraphs from Wikipedia pages, using candidate questions; 4) finally, we used crowd workers to evaluate each question and paragraph pair and provide the desired yes/no answers. We ensure high quality of the dataset by using a high overlap for annotation at the last step and a number of control gold-standard control questions, labelled by two of the authors. The core difference of DaNetQA to BoolQ is that some question may occur multiple times in the dataset, as at the step 3) we may retrieve more than one relevant paragraph. To make the dataset more challenging, we admit contradictory answers to a question if these answers are implied from the passages.









\subsubsection{Statistics for the Tasks}
Table \ref{table:stats} below presents the characteristics of the collected datasets - examples partitioning by train/val/test, as well as the total volume in tokens and sentences. As one can see, the size of the RuCoS task significantly exceeds the rest of the tasks due to the articles included in the task. 

\begin{table}[tbh!]
\centering
\small
\begin{tabular}{|l|l|l|l|}
\hline
Task                                & Samples & Sents & Tokens  \\ \hline
LiDiRus                                                   & 0/0/1104                     & 2210                       & $3.6\cdot10^4$             \\ \hline
\multicolumn{4}{|c|}{Common Sense}    \\\hline
RUSSE                                                     & 19845/8508/12151             & 90862                      & $1.1\cdot10^6$                  \\
PARus                                                     & 500/100/400                  & 1000                       & $5.4\cdot10^3   $                \\ \hline
\multicolumn{4}{|c|}{NLI}                            \\\hline
TERRa                                                     & 2616/307/3198                & 13706                      & $2.53\cdot10^5$               \\
RCB                                                       & 438/220/348                  & 2715                       & $3.7\cdot10^4    $                \\ \hline
\multicolumn{4}{|c|}{Reasoning}                       \\\hline
RWSD                                                      & 606/204/154                  & 1541                       & $2.3\cdot10^3$                 \\ \hline
\multicolumn{4}{|c|}{Machine Reading}  \\\hline
MuSeRC                                                    & 500/100/322                  & 12805                      & $2.53 \cdot 10^5$                  \\
RuCoS                                                     & 72193/4370/4147              & 583930                     & $1.2\cdot10^7$             \\ \hline
\multicolumn{4}{|c|}{World Knowledge}                     \\\hline
DaNetQA                                                   & 392/295/295                  & 6231                       & $1.31\cdot10^5$                   \\ \hline
\end{tabular}
\caption{Cumulative task statistics. The size train/validation/test splits is  provided in ``Samples'' column.}
\label{table:stats}
\end{table}

\subsection{Scoring}
Following \cite{wang2019superglue}, we calculate scores for each of the tasks based on their individual metrics. All metrics are scaled by 100x (i.e., as percentages). These scores are then averaged to get the final score. For the tasks with multiple metrics, the metrics are averaged.

\section{Experiments} \label{sec:experiments}

\subsection{Baselines}
In this section, we provide a two-step baseline design. At first we have developed a naïve baseline based on the TF-IDF model (section~\ref{sec:simple-baseline}), and then evaluate state-of-the-art models for Russian language (section~\ref{sec:advanced-baseline}).  

\subsubsection{Naïve Baseline}
\label{sec:simple-baseline}
We used Scikit-learn package~\cite{pedregosa2011scikit} to train a TF-IDF model. We used a 20 thousand sample from Wikipedia, from Russian and English sites equally. We restricted a vocabulary to 10 thousand most common words.
Then for each task set a logistic regression was trained to predict an answer. 

\begin{table*}[!htbp]
\centering
\begin{tabular}{|l|l|l|l|l|l|}
\hline
\textbf{Dataset}   & \textbf{Metrics} & \texttt{RuBERT}      & \texttt{MultiBERT}   & TF-IDF & Human                           \\  \hline
LiDiRus & MCC     & 0.186       & 0.157       & 0.059   & \textbf{0.626}                           \\ 
RCB     & $F_1$/Acc. & 0.432/0.468 & 0.383/0.429 & 0.45   & \textbf{0.68/0.702  }                    \\ 
PARus   & Acc     & 0.61        & 0.588       & 0.48   & \textbf{0.982}                           \\ 
MuSeRC  & $F_1$/EM  & 0.656/0.256 & 0.626/0.253 & 0.589/0.244      & \textbf{0.806}/\textbf{0.42 }                     \\ 
TERRa   & Acc     & 0.639       & 0.62        & 0.47   & \textbf{0.92}                            \\ 
RUSSE   & Acc     & \textbf{0.894}       & 0.84        & 0.66   & 0.747                           \\ 
RWSD    & Acc     & 0.675       & 0.675       & 0.66   & \textbf{0.84 }                           \\ 
DaNetQA & Acc     & 0.749       & 0.79        & 0.68   & \textbf{0.879}                           \\ 
RuCoS   & $F_1$/EM   & 0.255/0.251 & 0.371/0.367 & 0.256/0.251   & \textbf{0.93}/\textbf{0.924} \\ \hline
\textit{Average}   &         & 0.546           & 0.542       & 0.461      & 0.802                               \\ \hline
\end{tabular}
\caption{Results of the human benchmark and the baseline models. MCC stands for Matthews Correlation Coefficient; Acc - Accuracy; EM - Exact Match.}
\label{table:results}
\end{table*}

\subsubsection{Advanced Baselines} 
\label{sec:advanced-baseline}

We leverage two BERT-derived models as baseline. Multilingual BERT (\texttt{MultiBERT}), released by \cite{devlin2019bert}, is a single language model pre-trained from monolingual corpora in 104 languages, Russian texts being a part of training data. \texttt{MultiBERT} uses a shared vocabulary for all languages. The capabilities of \texttt{MultiBERT} for  zero-shot cross-lingual tasks have been recently studied by \cite{pires2019multilingual}. Russian BERT (\texttt{RuBERT}) was trained on large-scale corpus of news and Wikipedia in Russian. To alleviate the training all weights except sub-word embeddings were borrowed from \texttt{MultiBERT}. The sub-word vocabulary was obtained from the same training corpus and the new mono-lingual embeddings were transformed from the multi-lingual ones. This allowed to incorporate longer Russian sub-word units into the vocabulary. This model is  part of DeepPavlov framework \cite{kuratov2019adaptation}. 

\subsection{Human Evaluation} 

We include human performance estimates for all provided benchmark tasks, including the diagnostic set. We estimate human performance by hiring crowd workers via Toloka platform to re-annotate a sample from each task test set. We suggest a two step procedure: 1) a crowd worker is provided with an instruction and completes a short training phase before proceeding to the annotation phase, 2) a crowd worker that passed through the training phase solves the original test set.

For the annotation phase we ask crowd workers to annotate the full test sets except for the RUSSE and the RuCoS datasets, where we randomly sampled only 5000 and 1000 examples from the tasks' test sets, respectively. For each sample, we collect annotations from three to five crowd workers and take a majority vote to estimate human performance. In annotation phase we add control questions to prevent the crowd workers from cheating. As a result, we reject the annotations from the crowd workers that fail the training phase and do not include the results of those who achieved low performance on the control tasks.
The results of human evaluation are presented in Table~\ref{table:results}. The example of a Toloka task is provided in Appendix.

\section{Results}

The analysis of Table~\ref{table:results} can give an exact representation of the baseline model performance, which still remains significantly different from the human level. Nevertheless, the task of resolving the ambiguity of the word meaning in context (RUSSE) was solved by both monolingual and multilingual BERT at a level significantly exceeding the human one (0.89 vs 0.74). Besides, the monolingual model is showing a slightly higher quality than that of the multilingual one, especially prevailing textual entailment tasks (RCB, TERRA, PARus), disambiguating word meaning (RUSSE) and reading comprehension (MuSeRC). The multilingual model shows the most excellent result on the smallest dataset on commonsense QA task (DaNetQA) and also on commonsense-related task on machine reading (RuCoS).

We hope that our benchmark will help to excel the performance of models for the Russian language in the future, and will favour achieving comparably high results.

Can the results of a multilingual BERT on Russian and English data be considered analogous? Based on the results of the assessment, \texttt{MultiBERT} in English gets an overall score of 60.8 \footnote{Jiant, full SuperGLUE task set}, while on RussianGLUE task set an overall score of 54.2 is achieved-- 6\% lower, but noting that the English benchmark includes additionally Winograd Gender Parity~\cite{levesque2012winograd} dataset, giving SOTA models from 90 to 93\% of accuracy added to the overall assessment.  In the next section, a detailed comparison of the multilingual model performance is provided.

\subsection{Comparison to SuperGLUE}
As mentioned in Section~\ref{sec:tasks}, the diagnostic dataset has been obtained by professional translation with preservation of the original linguistic features mentioned.
Thus being said, this diagnostic data is the first of its kind that allows drawing a multilingual analogy of comparable models.

\begin{figure}[!htbp]
\centering
\includegraphics[scale=.5]{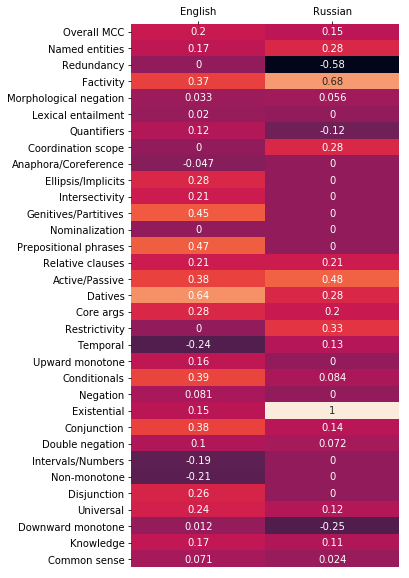}
\caption{Russian and English Diagnostic Evaluation on Multilingual transformer, scored using
Matthews’ correlation (MCC).}
\label{fig:diag-res}
\end{figure}

Procedure: using the original \texttt{MultiBERT}~\cite{devlin2019bert}, we conducted sequential model pretraining in English and Russian using the RTE dataset, and then tested the models on the diagnostic set, as long as the task requires exactly the same format. Predictions were further scored using Matthews’ correlation (MCC), and correlation for different linguistic features was computed. The results are presented in Figure~\ref{fig:diag-res}.

First of all, it could be noticed that the English variant of the model performs slightly better and shows a higher overall correlation of $0.2$ compared to $0.15$ for the Russian variant. This could be due to an asymmetry of the quality of the multilingual model and its better understanding of the English language in general.

As for the models' performance in the context of different linguistic features, the results generally coincide. For those categories for which correlation is low in English, the result in Russian is in most cases poor as well (for instance, \textit{Redundancy, Nominalization, Intervals/Numbers}). However, there exist several categories which are much better solved in English than in Russian such as \textit{PA Structure, Ellipsis/Implicits, Genetives/Partitives, Prepositional phrases, Datives} -- mostly low-level and/or syntactically driven categories, that may indicate that optimal hyperparameters of BERT architecture are much more suitable for English syntax and may not be linguistically universal. Similarly, we could find categories which show an extremely high correlation in Russian and low correlation in English (\textit{Factivity, Coordination scope, Restrictivity} and \textit{Existential}) -- high-level logical and semantic categories. 

These numbers compared to the ones for English could be explained by the fact that the language features now included in the diagnostics are not exactly linguistically universal in different languages and are mostly focused on the English language (at least those syntactic ones). Thus, for the comprehensive cross-linguistic typological analysis of possible linguistic features should be reviewed.

\section{Discussion} \label{sec:discussion}

We hope that our project will give a start to new research in the application of universal language models and transformers, including multilingual ones. Our example of an analysis of translated diagnostics shows that even in languages of the same European family (which Russian and English belong to), significant differences in the influence of linguistic categories on model performance are possible. One of the directions of the next studies, we consider detailed experiments on the influence of model parameters and language categories in data on the quality of the model in different languages.

An independent problem for the English original leaderboard is that a gradual improvement in the quality of models allows us to exceed the human performance level in individual tasks, as happened with the T5 \cite{t5} model. We expect that a similar situation will soon happen on Russian data, which means that when releasing straight off with complex SuperGLUE tasks, we will still be focused on adding tasks of a higher level of complexity in the future. Such tasks can become those that are obviously inaccessible to models for the ``understanding'' of long texts and documents, seq2seq tasks, tasks that require knowledge graphs.

In the further development of our leaderboard, we also see the possibility of adding an industrial assessment of models: for fair ranking and ease of use, all models could receive an estimate of the required memory resources, an estimate of performance, and so on.

\section{Conclusion} \label{sec:conclusion}
In this paper we present the first benchmark on general language understanding evaluation for the Russian language. The benchmark including nine task sets is aimed to test BERT-like models for their ability to perform entailment recognition, commonsense reasoning and machine reading while denoising various linguistic features added on the level of semantics, logical and syntactic structure.


We invite developers, researchers, and AI experts to join our project. 
Further development of the benchmark includes areas such as evaluation of industrial performance of models on the leaderboard and multilingual diagnostics.

\section*{Acknowledgements}
Ekaterina Artemova works within the framework of the HSE University Basic Research Program and funded by the Russian Academic Excellence Project  ``5-100''.

\bibliography{anthology,emnlp2020,ml}
\bibliographystyle{acl_natbib}

\appendix

\section{Appendices}

\includegraphics[width=6cm, height=3cm]{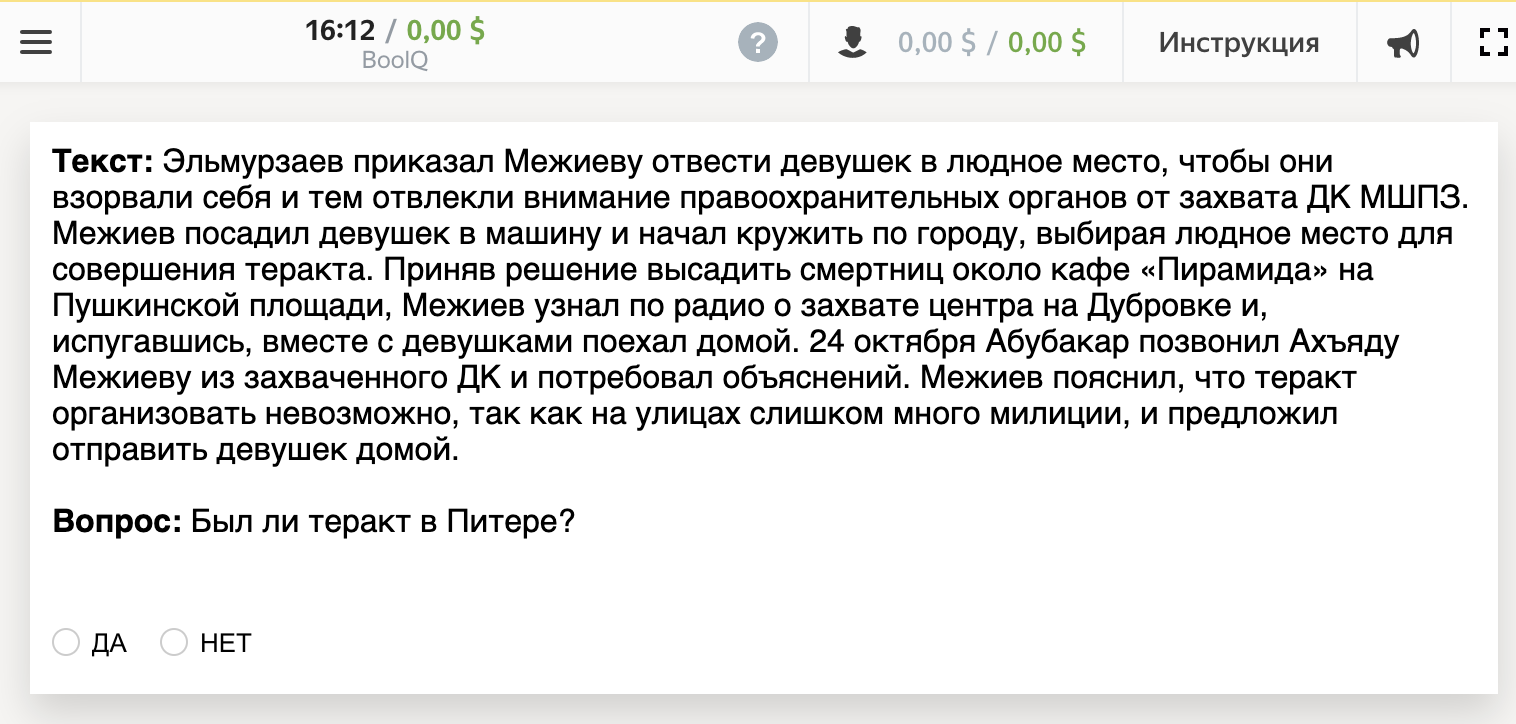}

\newpage

\section{Examples}

\begin{figure}[!htp]
\includegraphics[width = 0.45 \textwidth]{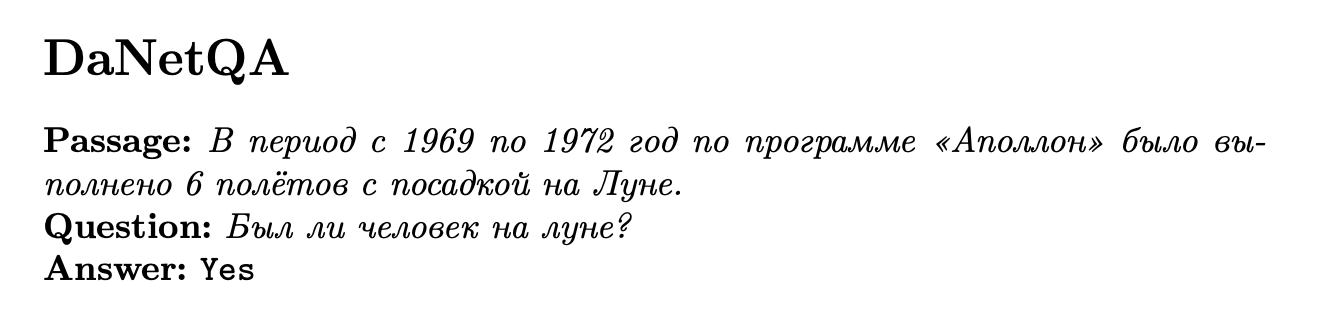}

\includegraphics[width = 0.45 \textwidth]{examples/muserc.png}

\includegraphics[width = 0.45 \textwidth]{examples/parus.png}

\includegraphics[width = 0.45 \textwidth]{examples/rcb.png}

\includegraphics[width = 0.45 \textwidth]{examples/rucos.png}

\includegraphics[width = 0.45 \textwidth]{examples/rwsd.png}

\includegraphics[width = 0.45 \textwidth]{examples/russe.png}

\includegraphics[width = 0.45 \textwidth]{examples/terra.png}

\end{figure}

\end{document}